\documentclass[11pt,a4paper]{article} 
\usepackage[hyperref]{naaclhlt2018}
\usepackage{times} 
\usepackage{latexsym}
\usepackage{url}
\usepackage{amsfonts}
\usepackage{amsmath}
\usepackage{amssymb}
\usepackage{graphicx}
\usepackage{stmaryrd}
\usepackage[shortlabels]{enumitem}
\usepackage{lipsum}
\usepackage{booktabs}

\newcommand\FigTop[4]{\begin{figure}[t] \begin{center} \includegraphics[scale=#2]{#1} \end{center} \caption{\label{fig:#3} #4} \end{figure}}

\newcommand\reffig[1]{Figure~\ref{fig:#1}}

\aclfinalcopy 
\def\aclpaperid{} 

\newcommand\nl[1]{{\it``#1''}}

\newcommand\compwebq{\textsc{ComplexWebQuestions}}
\newcommand\webq{\textsc{WebQuestions}}

\newcommand\simpqa{\textsc{SimpQA}}
\newcommand\splitqa{\textsc{SplitQA}}
\newcommand\docqa{\textsc{DocumentQA}}

\newcommand\ignore[1]{}

\title{Repartitioning of the \textsc{ComplexWebQuestions} Dataset}

\author{Alon Talmor \\ Tel-Aviv University \\ {\small \tt alontalmor@mail.tau.ac.il} \\\And Jonathan Berant \\ Tel-Aviv University \\ {\small \tt joberant@cs.tau.ac.il} \\}

\date{}

\begin{document} 
\maketitle 
\begin{abstract}  
Recently, \newcite{talmor2018web} introduced \textsc{ComplexWebQuestions} -- a dataset focused on answering complex questions by decomposing them into a sequence of simpler questions and extracting the answer from retrieved web snippets. In their work the authors used a pre-trained reading comprehension (RC) model \cite{salant2018contextualized} to extract the answer from the web snippets. In this short note we show that training a RC model directly on the training data of \textsc{ComplexWebQuestions} reveals a leakage from the training set to the test set that allows to obtain unreasonably high performance. As a solution, we construct a new partitioning of \textsc{ComplexWebQuestions} that does not suffer from this leakage and publicly release it. We also perform an empirical evaluation on these two datasets and show that training a RC model on the training data substantially improves state-of-the-art performance.
\end{abstract}

\section{\textsc{ComplexWebQuestions} Dataset} \label{sec:intro}

\textsc{ComplexWebQuestions} is a recently introduced Question Answering (QA) dataset \cite{talmor2018web}. Each example in \compwebq{} is a triple $(q, a, D)$, where $q$ is a question, $a$ is the correct answer (with aliases) and $D$ is a document, which contains a list of web snippets retrieved by a base model while attempting to answer the question. The dataset can be used by interacting with a search engine to find web snippets, or by using the pre-retrieved snippets. Table~\ref{tab:example_questions} provides a few examples for questions from the dataset.

\begin{table}[h]
\begin{center}
\scriptsize{
\begin{tabular}{l}
 \toprule
 \textbf{Question} \\ 
 \midrule
\nl{What films star Taylor Lautner and have costume designs by Nina Proctor?}  \\ 
\nl{Which school that Sir Ernest Rutherford attended has the latest founding date?} \\
\nl{Which of the countries bordering Mexico have an army size of less than 1050?}\\
\nl{Where is the end of the river that originates in Shannon Pot?}\\ 
\toprule
\end{tabular}}
\end{center}
\caption{Example questions from \compwebq{}.}
\label{tab:example_questions}
\end{table}

\compwebq{} was created by taking examples from the dataset \textsc{WebQuestionsSP} \cite{yih2016value}, which contains 4,737 questions paired with SPARQL queries for Freebase \cite{bollacker2008freebase}.
In \webq{}, questions are broad but simple. Thus, the authors sampled question-query pairs, automatically created more complex SPARQL queries with manually-defined rules, generated automatically questions that are understandable to Amazon Mechanical Turk workers, and then had them paraphrased into natural language (similar to \newcite{wang2015overnight}). They computed answers by executing complex SPARQL queries against Freebase, and obtained broad and complex questions. \reffig{ComplexQuestionGeneration} provides an overview of this procedure. 

\FigTop{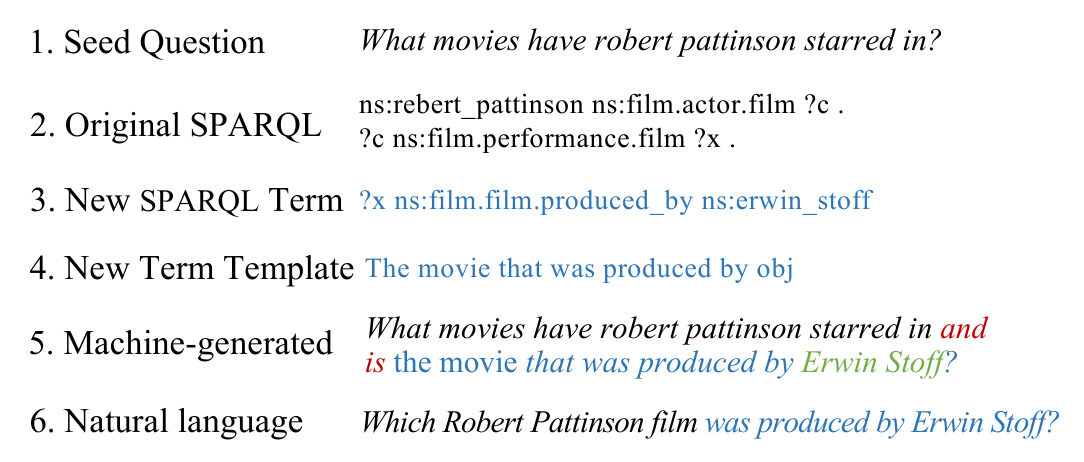}{0.77}{ComplexQuestionGeneration}{Overview of data collection procedure. Blue text denotes different stages of the term addition, green represents the obj value, and red the intermediate text to connect the new term and seed question.}

\section{Partitioning Issue}

\begin{table*}[t]
\begin{center}
\scriptsize{
\begin{tabular}{l|l}
\toprule
\textbf{Complex Question}  & \textbf{Answer} \\ 
\midrule
\nl{Who was the president in 1980 of the nation that uses the Pakistani repuu as money?} & \nl{Muhammad Zia-ul-Haq}  \\
\nl{The country that contains Balochistan, Pakistan had what President in 1980?} & \nl{Muhammad Zia-ul-Haq} \\
\nl{Who held the office of president in 1980 in the country that has Islamabad as its capital?} & \nl{Muhammad Zia-ul-Haq}\\
\toprule
\end{tabular}}
\end{center}
\caption{Example of a seed question \nl{Who was the president of Pakistan in 1980?} and three questions that were generated from it.}
\label{tab:spurious_example}
\end{table*}

The original \compwebq{} was created by generating 34,689 examples and then randomly partitioning them into disjoint training, development and test sets. Because every seed question from \webq{} produces multiple questions in \compwebq{}, questions that originate from the same seed question will be placed in both the training set and test set with high probability. Consider the example in Table~\ref{tab:spurious_example} of a seed question and three questions that were generated from it.

As can be seen in this example, if one of the generated questions is placed in the training set and another in the test set, the model can learn a spurious correlation between the terms \nl{1980} and \nl{Pakistan} to \nl{Muhammad Zia-ul-Haq}. Naturally, this is not a desired behavior.

The model presented by \newcite{talmor2018web} did not train a RC model on \compwebq{} and instead used a model that was pre-trained on the SQuAD dataset \cite{rajpurkar2016squad}. Thus, their model did not learn any of the spurious correlations mentioned above. In this work, we train a RC directly on \compwebq{} and find that the model indeed learns these correlations and thus spurious information is leaked from the training set to the test set.

The solution is simple -- we repartition the \compwebq{} dataset based on the seed questions, that is, we randomly split the original \webq{} questions into a training, development and test set. Thus, no question in the test set (or development set) originates from the same seed question as a question from the training set. We train a RC model on this new dataset and show that its performance is substantially lower compared to the original partitioning. Nevertheless, training a RC model on \compwebq{} improves performance compared to the pre-trained model and thus we establish a new state-of-the-art on \compwebq.

We name the new dataset \compwebq{} version 1.1 (and the old dataset version 1.0), and it can be downloaded from \url{https://www.tau-nlp.org/compwebq}.





\section{Experiments} \label{sec:experiments}

In this section we show how training a RC model on \compwebq{} results in high performance on \compwebq version 1.0, and much lower performance on \compwebq version 1.1. As a side effect, we also report experiments on different ways of training the RC model on \compwebq{}.


\paragraph{Experimental setup}
The QA model of \newcite{talmor2018web} has two parts. First, a question decomposition model determines how to decompose the complex question into a sequence of simpler questions. Second, each question is sent to a search engine, web snippets are retrieved, and a RC model extracts the answer from the snippets. Thus the model is specified by (i) a question decomposition procedure (ii) a RC model. \newcite{talmor2018web} propose two decomposition procedures, which we also evaluate.
\begin{enumerate}
\item \simpqa: A question is not decomposed but sent as is to the search engine.
\item \splitqa: A question decomposition model decomposes the question and then re-composes the final answer.
\end{enumerate}

\newcite{talmor2018web} used a single RC model in their work that was pre-trained on \textsc{SQuAD}. Here, we evaluate three RC models
\begin{enumerate}
\item \textsc{Pretained}: The same model from \newcite{talmor2018web}.
\item \textsc{NoDecomp}: We train the \docqa{} RC model \cite{clark2017simple} on all question-answer pairs from \compwebq, where for each question we provide Google web snippets when using the entire question. The total number of examples \docqa{} is trained on is 24,649.
\item \textsc{Decomp}: We train \docqa{} on all examples from \textsc{NoDecomp}. However, due to the generation process of \compwebq{}, \newcite{talmor2018web} presented a method for heuristically decomposing complex questions into two simpler questions for which the answer can be computed from Freebase. Thus, we can train \docqa{} not only on the original complex questions, but also on the decomposed questions by sending them to a search engine and retrieving the snippets. This is important for \splitqa{}, which applies the RC model on simple rather than complex questions. In total, we train \docqa{} on 63,263 examples, derived from the \compwebq{} training set.
\end{enumerate}

For evaluation, we measure precision@1, the official evaluation metric of \compwebq{}. We now present empirical results for various combinations of a questions decomposition model and a RC model.

\subsection{Results}

\begin{table}[t]
\begin{center}
\footnotesize{
\begin{tabular}{l|c|c}
\textbf{System} & \textbf{Dev.} & \textbf{Test} \\
\hline
\simpqa+\textsc{Pretrained} & 20.4 & 20.8 \\
\splitqa+\textsc{Pretrained} & 29.0 & 27.5 \\
\hline
\simpqa+\textsc{NoDecomp} & 47.8 & - \\
\splitqa+\textsc{NoDecomp} & 55.0 & -\\
\hline
  
\end{tabular}}
\end{center}
\caption{precision@1 results on the development set and test set for \compwebq{} version 1.0}
\label{tab:1.0}
\end{table}

\begin{table}[t]
\begin{center}
\footnotesize{
\begin{tabular}{l|c|c}
\textbf{System} & \textbf{Dev.} & \textbf{Test} \\
\hline
\simpqa+\textsc{Pretrained} & 20.5 & 19.9 \\
\splitqa+\textsc{Pretrained}  & 27.6 & 25.9 \\
\hline
\simpqa+\textsc{NoDecomp} & 30.6 & - \\
\splitqa+\textsc{NoDecomp} & 31.1 & -\\
\splitqa+\textsc{Decomp}& 35.6 & 34.2 \\
\hline
  
\end{tabular}}
\end{center}
\caption{precision@1 results on the development set and test set for \compwebq{} version 1.1}
\label{tab:1.1}
\end{table}

Tables~\ref{tab:1.0} and \ref{tab:1.1} present the results of our evaluation on both versions of \compwebq{} and various models.

Results of \simpqa+\textsc{NoDecomp} and \splitqa+\textsc{NoDecomp} on version 1.0 show a significant increase in accuracy on the development set of 47.8 and 55.0 respectively. However, when comparing results of these models when trained on version 1.1, results are much lower -- 30.6 and 31.1 respectively. Conversely, \simpqa+\textsc{Pretrained}, that has not been trained on these training sets, remains uneffected by the repartitioning, and retains a similar precision@1 of 20.5 and 20.4. This demonstrates that version 1.0 did in fact enable the model to learn spurious correlations to achieve a much higher accuracy whereas in version 1.1 this is not the case.


Overall, results of \simpqa+\textsc{NoDecomp} and \splitqa+\textsc{NoDecomp} are higher compared to the pre-trained RC model, showing that training a RC model on the training set improves performance.
A further increase in accuracy in \splitqa{} from 31.1 to 35.6 on the development set is achieved by training the model on the full questions as well as decomposed questions as shown in \splitqa+\textsc{Decomp}, implying that adding the decomposed questions improves performance as well.
Finally, we run \splitqa+\textsc{Decomp} once on the test set, to achieve precision@1 score of 34.2, establishing a new state-of-the-art on \compwebq{}.

\section{Conclusion}

In this short note we describe a problem with the partitioning of the \compwebq{} dataset, which was exposed by training a RC model on this dataset. We solve this problem by re-partitioning the dataset, present an empirical evaluation, and report a new state-of-the-art on \compwebq{}. The new dataset is publicly available on the \compwebq{} website.
 
\bibliography{all}
\bibliographystyle{acl_natbib}
\end{document}